\documentclass[lettersize,journal]{IEEEtran}
\usepackage{amsmath,amsfonts}
\usepackage{algorithmic}
\usepackage{algorithm}
\usepackage{array}
\usepackage[caption=false,font=normalsize,labelfont=sf,textfont=sf]{subfig}
\usepackage{textcomp}
\usepackage{stfloats}
\usepackage{url}
\usepackage{verbatim}
\usepackage{graphicx}
\usepackage{cite}
\usepackage{xcolor}
\usepackage{multirow}
\usepackage{colortbl}
\usepackage{hyperref}

\begin{document}

\title{2D-ThermAl: Physics-Informed Framework for \underline{Therm}a\underline{l} Analysis of Circuits using Generative \underline{A}I}

\author{Soumyadeep Chandra, Sayeed Shafayet Chowdhury, and Kaushik Roy~\IEEEmembership{Fellow,~IEEE}
\thanks{The authors are with the Department of Electrical and Computer Engineering, Purdue University, West Lafayette, IN, USA - 47906 (e-mail: chand133@purdue.edu, chowdh23@purdue.edu, kaushik@purdue.edu). 

This work was supported in part by the Center for Co-design of Cognitive Systems (CoCoSys), one of the seven centers in JUMP 2.0, a Semiconductor Research Corporation (SRC) program sponsored by DARPA, by the SRC, the National Science Foundation, Intel Corporation, the DoD Vannevar Bush Fellowship, and by the U.S. Army Research Laboratory. All authors declare that they have no conflicts of interest. 

Corresponding author: Soumyadeep Chandra (e-mail: chand133@purdue.edu)}
\thanks{Manuscript received April 19, 2021; revised August 16, 2021.}}

\markboth{Journal of \LaTeX\ Class Files,~Vol.~14, No.~8, August~2021}%
{Shell \MakeLowercase{\textit{et al.}}: A Sample Article Using IEEEtran.cls for IEEE Journals}


\maketitle

\begin{abstract}
Thermal analysis is increasingly critical in modern integrated circuits, where non-uniform power dissipation and high transistor densities can cause rapid temperature spikes and reliability concerns. Traditional methods such as FEM-based simulations offer high accuracy but computationally prohibitive for early-stage design, often requiring multiple iterative redesign cycles to resolve late-stage thermal failures. To address these challenges, we propose \textit{`ThermAl'}, a physics-informed generative AI framework which effectively identifies heat sources and estimates full-chip transient and steady-state thermal distributions directly from input activity profiles. ThermAl employs a hybrid U-Net architecture enhanced with positional encoding and a Boltzmann regularizer to maintain physical fidelity. Our model is trained on an extensive dataset of heat dissipation maps for more than 200 circuit configurations, ranging from simple logic gates (e.g., inverters, NAND, XOR) to complex designs, generated via COMSOL and Cadence EDA flows. The dataset captures diverse activity patterns, and we note that material-dependent thermal properties may require targeted fine-tuning to ensure accuracy across different fabrication contexts. Experimental results demonstrate that ThermAl delivers precise temperature mappings for large circuits, with a root mean squared error (RMSE) of only $0.71^{\circ}$C and outperforms conventional FEM tools by running up to $\sim200\times$ faster. We analyze performance across diverse layouts and workloads, and discuss its applicability to large-scale EDA workflows. While thermal reliability assessments often extend beyond 85$^{\circ}$C for post-layout signoff, our focus here is on early-stage hotspot detection and thermal pattern learning. To ensure generalization beyond the nominal operating range ($25-55^{\circ}$C), we additionally performed cross-validation on an extended dataset spanning $25-95^{\circ}$C maintaining a high accuracy ($<2.2\%$ full-scale RMSE) even under elevated temperature conditions representative of peak power and stress scenarios. Limitations such as 2D-only modeling and real-world validation are addressed with concrete future directions, including 3D extension, generalization across technology nodes, and transfer learning strategies. The code and dataset are publicly available at: \url{https://github.com/soumyadeepchandra/2D-ThermAl}
\end{abstract}


\section{Introduction}
\IEEEPARstart{T}{hermal} management has become a paramount concern in modern microprocessor design, as power densities continue to escalate with each generation of technology~\cite{taylor2013landscape,esmaeilzadeh2011dark}. The failure of Dennard scaling has dissolved the traditional transistor scaling–power balance, leading to chips that are increasingly constrained by thermal and reliability limits over pure performance~\cite{esmaeilzadeh2011dark}. To prevent overheating and thermal-induced failures, designers must integrate accurate temperature estimation and control mechanisms early in the design flow, enabling techniques such as clock gating, power gating, dynamic voltage and frequency scaling (DVFS), and task migration to operate effectively in real time~\cite{brooks2001dynamic,hanumaiah2012energy,liu2014task,wang2016hierarchical}. Current thermal‑analysis techniques fall into two groups: high‑fidelity finite‑element analysis (FEA) and coarse‑grain models. FEA tools capture detailed heat‑transfer physics but incur prohibitive runtimes and setup complexity for full‑chip, iterative exploration. In contrast, coarse methods—from analytical RC networks to infrared imaging—compromise spatial resolution, missing fine‑grain hotspots that drive localized failure modes~\cite{skadron2003temperature,kong2012recent}. These trade‐offs leave a pressing need for predictive methods that simultaneously deliver circuit‐level accuracy, computational efficiency, and seamless integration into early‑stage design.

\begin{figure}[!t]
    \centering
    \includegraphics[width=0.98\linewidth]{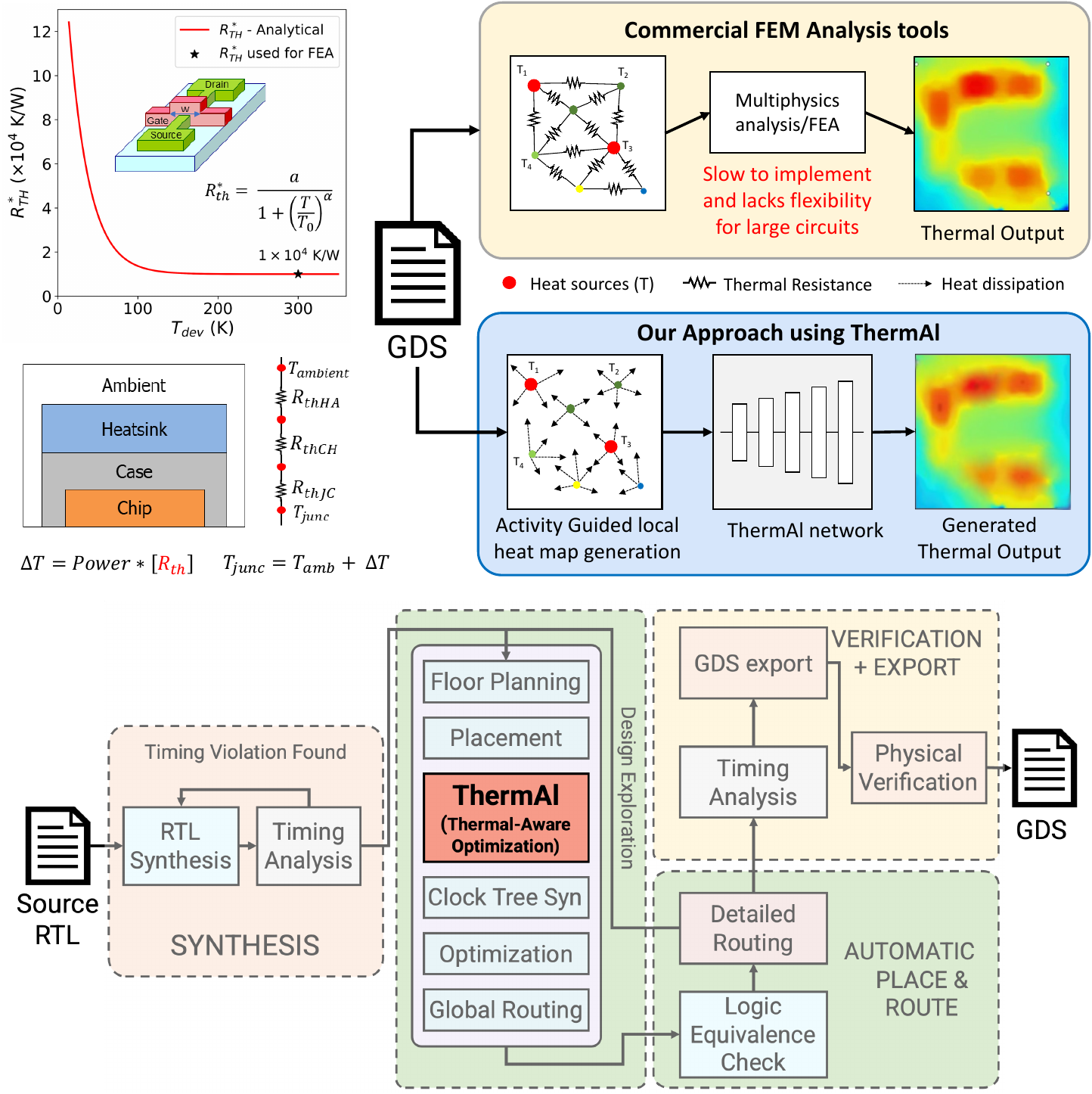}
    \caption{As transistor density in modern microprocessors increases, traditional thermal estimation methods often struggle with complex heat dissipation, leading to inaccuracies in temperature profiling. While commercial FEM tools offer high fidelity, they can be too slow and computationally inefficient for full-chip modeling. Simultaneously, thermal cameras tend to lack the high-resolution detail needed for analysis at the circuit level. ThermAl tackles these limitations by using a generative neural network to deliver fast and accurate thermal estimation. More importantly, it integrates easily into the design exploration stage of the GDS generation, allowing designers to identify and correct thermal hotspots early on. Thermal-sensitive flow can prevent numerous iterations of designs, induced by thermal failures discovered late in the process, and thus simplify the overall circuit optimization process.}
    \label{fig:motivation}
    \vspace{-1em}
\end{figure}

Although crucial, circuit-level thermal estimates are challenging to make due to complex heat transfer coupled with non-uniform patterns of power dissipation within current-day microprocessors. Early integration of thermal predictions aids designers in keeping thermal constraints within reach and eliminating expensive redesigns. Early identification of potential hotspots streamlines development. Traditional approaches based on core‑wise or functional‑unit power measurements lack the resolution to capture fine‑grained hotspot behavior~\cite{joseph2001run,isci2003runtime,wu2008efficient,dev2013power}. Inverse thermal‑map techniques recover power maps from thermal measurements via optimization and machine learning~\cite{wang2009power,nowroz2010thermal,reda2011improved,nowroz2012power,ranieri2012eigenmaps,paek2013powerfield,beneventi2015thermal,reda2017blind}, but they struggle to generalize at circuit‑level granularity when heat fluxes become highly localized. 

FEM commercial tools provide high fidelity, high computational cost, and difficulty setting them up render them unsuitable for large‑scale, real‑time applications, and even optimized methods like HotSpot~\cite{hotspot}, Power Blurring~\cite{powerblurring}, and PACT~\cite{PACT} grow costly at finer resolutions. Compact thermal models such as 3D-ICE~\cite{3DICE} and MTA~\cite{MTA} improve runtime efficiency for stacked 3D ICs and architecture-level exploration, but their grid coarseness limits spatial accuracy for dense sub-block thermal variations.

More recently, there have been more advanced learning-based methods. One approach utilizes spatial Laplace transforms to identify heat sources~\cite{sadiqbatcha2019hot}, while another employs GANs to predict full‑chip thermal maps from performance metrics~\cite{jin2020full}. Deep neural approaches like DeepOHeat~\cite{deepoheat} have attempted to infer steady-state temperature fields directly from power distributions, yet they often rely on data-driven correlations without explicit physical grounding, leading to limited generalization under new boundary conditions or materials. However, these remain reliant on imprecise, coarse-grained analysis due to limitations of thermal camera resolution or imperfect power‑thermal coupling. 

Moving beyond these challenges, we introduce a hybrid solution integrating detailed power estimation, circuit‑level boundary values, and sophisticated thermal simulation. We integrate equivalent thermal RC networks~\cite{lasance1995novel,gerstenmaier2002rigorous}, architecture‑level models~\cite{liu2006fast,li2009architecture}, and FEM methodologies~\cite{gurrum2008compact} (Fig.\ref{fig:motivation}) to accurately compute temperature gradients. We introduce \textbf{ThermAl}, a conditional generative framework that frames thermal analysis as an image‑to‑image translation task. The model generates steady‑state and transient thermal maps by conditioning on a sample pair of images $\{E, E^{\prime}\}$, capturing non‑uniform heat dissipation, material heterogeneity, and dynamic boundary conditions. 

Our dataset captures a wide spectrum of spatial and temporal activity patterns; however, the resulting thermal behavior is jointly determined by these activity patterns and the underlying material properties. Parameters such as thermal conductivity, specific heat capacity, and layer stack composition critically modulate how localized power dissipation evolves into spatial temperature distributions. Consequently, identical activity can yield substantially different hotspot intensities and diffusion profiles across substrates, interconnect materials, and thermal interface layers. This intrinsic coupling between activity-induced and material-dependent variability necessitates training strategies—such as targeted fine-tuning or transfer learning—that preserve ThermAl’s predictive accuracy and generalizability across diverse technology nodes and packaging configurations without compromising inference speed. While the current framework is primarily developed and validated within a moderate temperature range of $25-55^{\circ}$C, corresponding to typical early-design power densities, we further conduct cross-validation on an extended dataset spanning $25-95^{\circ}$C to assess model robustness under elevated thermal stress. As detailed in Section~\ref{cv_result}, \textit{ThermAl} maintains high predictive fidelity (full-scale RMSE $<2.2\%$) even across this wider range, demonstrating strong generalization to high-temperature scenarios representative of peak power and reliability conditions.

Beyond prior generative AI and data-driven thermal modeling approaches, \textit{ThermAl} explicitly incorporates physics-informed learning within its neural architecture, addressing key limitations observed in purely data-driven frameworks. A physics-aware regularizer guides the network to learn heat diffusion dynamics that adhere to fundamental physical laws, thereby improving accuracy in both transient and steady-state regimes—particularly near complex thermal boundaries and hotspot regions. Compared to traditional FEM‑based tools, \textit{ThermAl} infers hundreds of times faster while providing high accuracy to enable quick, real-time thermal decision-making. The model achieves high prediction accuracy, with an RMSE of only $0.71^{\circ}$C for a $256 \times 256$ resolution grid. Additionally, the model supports higher-resolution inputs, achieving an RMSE of $1.18^{\circ}$C for $512 \times 512$ thermal maps.

\noindent Our main contributions are as follows:
\begin{enumerate}
    \item Unlike commercial FEM tools, which tends to be excessively slow and memory-hungry at high resolutions, \textit{ThermAl} employs a generative machine-learning architecture integrated with physics-based constraints, enabling accurate thermal analysis. This framework achieves an RMSE of only $0.71^{\circ}$C, while decreasing inference times by nearly $200\times$ compared to classic FEM in both transient and steady-state heat problems. 
    \item We assemble an extensive database of heat dissipation maps for over 200+ circuit configurations, ranging from simple logic gates (inverters, NAND, XOR) to complex workloads. These maps are generated using off-the-shelf commercial finite element Multiphysics software (such as COMSOL) to provide an exhaustive range of physical and geometrical boundary conditions required for learning and verification.
    \item Developing proposed \textit{ThermAl} employs an advanced hybrid U-Net architecture augmented with positional embeddings, feature-level concatenation, and physics-informed regularization. These enhancements significantly improve both spatial consistency and prediction accuracy of transient and steady-state thermal maps. By directly integrating fundamental heat transfer principles into the training loss, ThermAl effectively extrapolates thermal diffusion patterns within complex designs, achieving fast and precise temperature mapping.
    \item To evaluate generalization beyond nominal operating conditions ($25$–$55^{\circ}$C), we perform extended-range cross-validation over a small dataset spanning $25$–$95^{\circ}$C. Results demonstrate that \textit{ThermAl} maintains high accuracy (full-scale RMSE $<2.2\%$) and strong spatial consistency, confirming its robustness across both early-stage design and elevated thermal stress scenarios.

\end{enumerate}

\noindent The remainder of this article is organized as follows: Section~\ref{related} summarizes state‑of‑the‑art thermal‑analysis methodology, emphasizing key principles and limitations. Section~\ref{dataset} provides a comprehensive overview of the dataset generation using commercial EDA and FEA tools. Section~\ref{methods} introduces our conditional generative machine-learning model, detailing architectural choices and training methodology. Section~\ref{experiment} describes the experimental system and performance evaluation metrics. Section~\ref{results} presents the inference speed, accuracy, and efficiency of the model for varied scenarios. Section~\ref{ablation} discusses an ablation study that validates key design choices, and Section~\ref{conclusion} summarizes principal findings and potential avenues for future work.

\section{Background and Related Work} \label{related}
\subsection{System‐Level Thermal Management}
System‑level schemes monitor package‑ or die‑level temperatures and react by throttling performance or migrating workload. Brooks et al.\cite{brooks2001dynamic} introduced thermal feedback-driven dynamic voltage/frequency scaling (DVFS), which reduced average junction temperatures but often overthrottled to avoid worst-case hotspots. Hanumaiah and Ranka\cite{hanumaiah2012energy} proposed proactive DVFS by predicting thermal trends based on performance counters, improving responsiveness but remaining limited by coarse-grained sensor data. Liu et al.\cite{liu2014task} and Wang et al.\cite{wang2016hierarchical} advanced these techniques with fine-grained task migration across cores and sockets, balancing load to flatten temperature peaks. However, these methods primarily rely on chip- or core-level averages, rendering them ineffective at detecting sub-circuit hotspots and thus limiting their capacity to prevent localized reliability degradation (e.g., electromigration or negative bias temperature instability).

\begin{figure}[!ht]
    \centering
    \includegraphics[width=\linewidth]{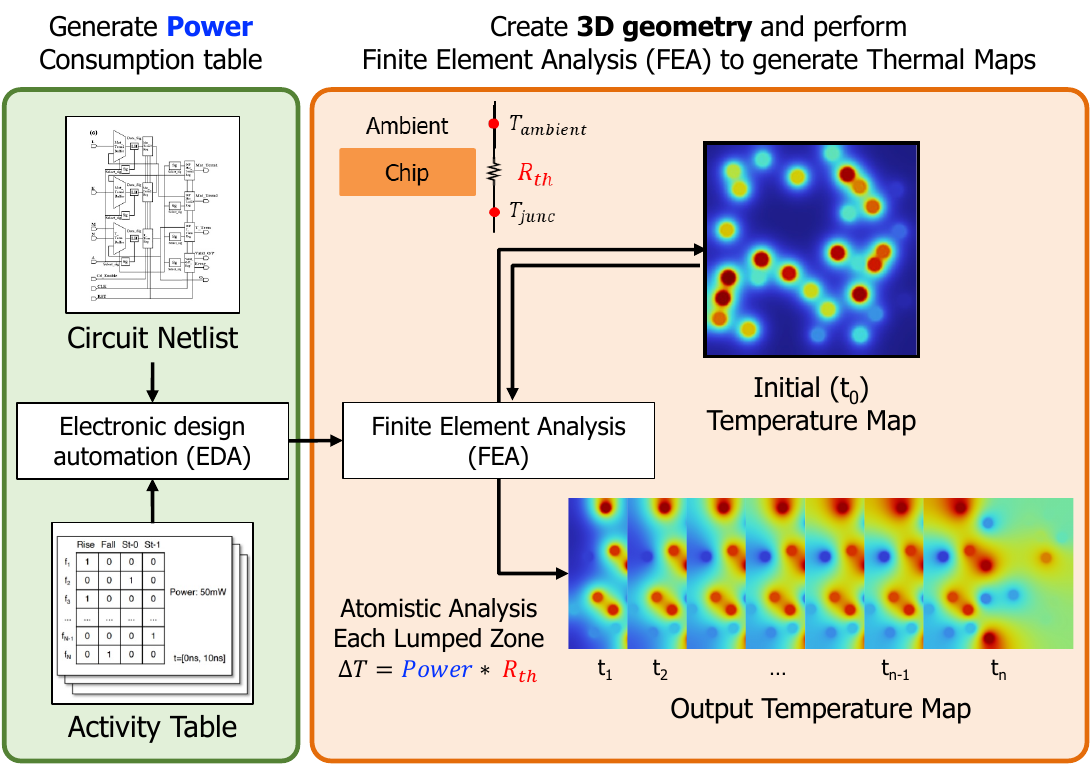}
    \vspace{-1em}
    \caption{Workflow for Dataset Generation: Circuit designs and power distribution data are first generated using standard EDA tools. These, along with specified thermal parameters, are input into FEM-based time-stepped simulations that capture transient heat flow, producing thermal maps at various intervals. The resulting comprehensive dataset serves as ground truth for training \textit{ThermAl}, enabling accurate predictions across diverse circuit configurations and thermal conditions.}
    \vspace{-1em}
    \label{fig:dataset}
\end{figure} 

\vspace{-0.5em}
\subsection{Inverse Thermal Mapping}
Inverse thermal mapping frames power reconstruction as an ill-posed optimization problem: estimating the underlying spatial power distribution given a temperature field. Wang et al.\cite{wang2009power} applied conjugate-gradient inversion with Tikhonov regularization, recovering coarse power maps but suffering from smoothing artifacts. Nowroz et al.\cite{nowroz2010thermal,nowroz2012power} improved stability through multi-resolution wavelet bases and edge-preserving priors, improving hotspot localization but necessitating dense sensor arrays. Reda et al.\cite{reda2011improved,reda2017blind} developed blind deconvolution techniques using learned point-spread functions, enabling power recovery from sparse measurements but still constrained by solution non-uniqueness. Ranieri et al.\cite{ranieri2012eigenmaps} and Paek et al.~\cite{paek2013powerfield} proposed dimensionality reduction via eigenbasis representations, simplifying inversion complexity but limiting adaptability to arbitrary layouts. 
Collectively, these approaches demonstrate the potential of inverse methods for power profiling but struggle with localized, circuit‑level granularity and measurement noise.

\vspace{-0.5em}
\subsection{Thermal-modeling Frameworks}
Analytical models such as HotSpot~\cite{hotspot} conceptualize the die as an RC network for rapid temperature estimation but require manual grid tuning and struggle with fine-grained, sub-circuit accuracy, limiting their utility for highly dense modern layouts. Power Blurring~\cite{powerblurring} improves spatial precision by convolving detailed power maps with precomputed impulse response kernels, though it introduces significant setup complexity and limited flexibility under dynamic workloads. PACT~\cite{PACT} offers a parameterizable thermal modeling engine that decomposes the die into reusable thermal units calibrated offline, enabling rapid, workload‑aware estimation while still demanding significant preprocessing and facing challenges in scaling to highly heterogeneous layouts. Although these methods reduce runtime compared to full finite-element analysis (FEA), their reliance on manual tuning and complex configuration hinders seamless integration into contemporary electronic design automation (EDA) workflows. More comprehensive solvers like 3D-ICE~\cite{3DICE} and the Manchester Thermal Analyzer (MTA)~\cite{MTA} explicitly couple chip, interconnect, and heat-sink domains to model full 3D thermal interactions, reporting speedups of up to $975\times$ and $>100\times$ respectively over conventional FEM tools. However, their reliance on detailed package-level characterization, extensive preprocessing, and solver-based numerical integration often restricts their use during early design exploration.

\vspace{-0.5em}
\subsection{Learning‑Based Thermal Estimation}
Deep learning has recently been applied to directly map performance counters or low-resolution thermal inputs to full-chip thermal fields. Sadiqbatcha et al.~\cite{sadiqbatcha2019hot} introduced a CNN leveraging Laplacian spectral features to improve hotspot detection beyond analytical models, but lacking transient‐state predictions. Jin et al.~\cite{jin2020full} trained a conditional GAN on synchronized performance data and infrared camera outputs to synthesize realistic steady-state and transient maps; however, reliance on external imaging hardware limited resolution to millimeter scales, hereby constraining practical scalability. More recently, Chen et al.~\cite{chen2021deep} utilized a UNet architecture to regress high-resolution thermal maps from simulated power traces, demonstrating subpixel accuracy in synthetic benchmarks. Transformer-based architectures~\cite{wang2022transformthermal} have also been explored to capture long-range heat diffusion patterns, though they incur high training costs and require extensive datasets to generalize across layouts. Recent operator-learning approaches such as DeepOHeat~\cite{deepoheat} achieve $10^3–10^5\times$ speedups with sub-degree accuracy, but rely on large-scale 3D datasets, long convergence times, and high memory overhead.

By synthesizing insights from system-level controls, inverse mapping, learning-based inference, and hybrid modeling strategies, our proposed \textit{ThermAl} framework adopts a balanced strategy—embedding physics-based priors within a generative U-Net to maintain physical consistency while preserving the low-latency benefits of learning-based inference. It directly learns circuit-scale thermal dynamics across diverse boundary conditions and delivers near-real-time predictions without requiring a customized grid setup or external hardware. This positions ThermAl between analytical compact solvers and heavy operator-learning models, offering a tractable, layout-aware framework for rapid thermal estimation during early EDA design stages.

\section{Dataset Generation} \label{dataset}
The construction of a comprehensive and diverse dataset is a cornerstone for effectively training our generative AI framework for thermal analysis. This dataset supports the ability of the model to generalize and predict thermal behavior in a wide variety of circuit designs and operational configurations. In this work, we generate over 200 synthetic thermal maps representing a broad spectrum of circuit configurations. These include simple combinational elements, such as inverters, NAND, and XOR gates; mid-level designs like adders and multiplexers; and complex functional blocks as summarized in Table~\ref{tab:circuit-benchmarks}. Each configuration was simulated under multiple power profiles and boundary conditions to ensure coverage of practical thermal scenarios. Such breadth ensures comprehensive coverage of operational scenarios essential for both training and inference.

\begin{table}[ht]
\centering
\caption{Overview of Circuit Benchmark Configurations}
\resizebox{\linewidth}{!}{%
\begin{tabular}{|c|c|c|c|}
\hline
\textbf{Circuit Type} & \textbf{Examples} & \textbf{Instances} & \textbf{Gate Count} \\
\hline
Basic Logic Gates & Inverter, NAND, XOR & 120 & $<$ 50 \\
\hline
Sequential Elements & Flip-Flops, Latches & 60 & $\sim$ 50--100 \\
\hline
Combinational Blocks & Adder, Multiplexer & 40 & $\sim$ 100--400 \\
\hline
\end{tabular}%
}
\label{tab:circuit-benchmarks}
\end{table}

\begin{figure}[!ht]
    \centering
    \includegraphics[width=\linewidth]{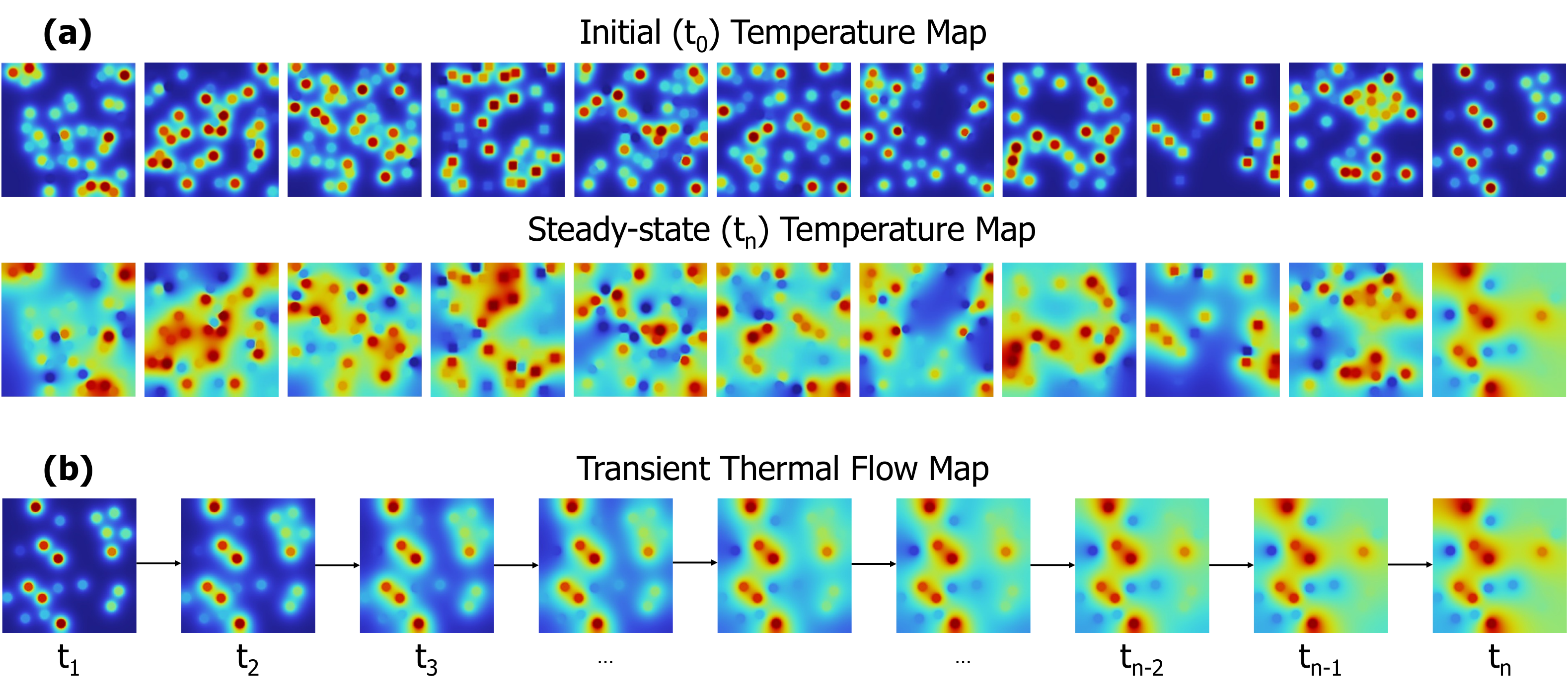}
    \vspace{-2em}
    \caption{(a) Comprehensive library of steady-state and transient thermal behaviors across diverse workloads and initial conditions, generated using the commercial finite element Multiphysics software (COMSOL). (b) Transient snapshots illustrate the time-evolving evolution of heat flow, highlighting changes in temperature profiles over time within the circuit.
    }
    \label{fig:data}
\end{figure}

To generate such a varied dataset, we employ Electronic Design Automation (EDA) tools such as \textit{CADENCE}\cite{cadence2024}, which provide detailed insights into power distribution and transistor activity under varied operating conditions (Fig.~\ref{fig:data}(a)). Standard cell libraries, such as the TSMC 65nm~\cite {tsmc65nm}, serve as the foundational design space, capturing key electrical characteristics, including netlists and extensive power profiles, across multiple scenarios. By pairing this electrical data with thermal parameters ($R_{th} = 1\times10^4$ K/W) outlined in~\cite{casse2022low}, we capture the microscopic interplay between material properties and heat flow dynamics. The current setup assumes constant thermal resistance at the die-ambient interface, effectively modeling air-cooled conditions. While suitable for algorithmic benchmarking, this simplification omits heat sink dynamics and varying boundary conditions (e.g., water cooling). Extending ThermAl to include coupled package–sink modeling is a planned direction for future work.

We integrate the electrical data into finite element  methods (FEM) using COMSOL Multiphysics® software~\cite{comsol} to generate detailed thermal simulations for each circuit configuration, as shown in Fig.~\ref{fig:dataset}. Each thermal image consists of a $256\times256$ pixel grid, mapping to a die area of $1,\text{mm} \times 1,\text{mm}$ and assigning each pixel to approximately $3.9,\mu\text{m} \times 3.9,\mu\text{m}$ of physical silicon. The pixel pitch of $3.9,\mu\text{m}$ aligns well with the characteristic thermal diffusion length in silicon over microsecond time scales. Using a typical thermal diffusivity $(\alpha \approx 1\times10^{-4} \text{m}^2/\text{s})$, the diffusion length $L = \sqrt{\alpha t}$ is about $15$–$20\mu\text{m}$, which indicates that heat from a localized hotspot spreads over roughly this range. As each pixel covers $3.9\mu\text{m}$, this spread spans around 3–5 pixels, ensuring that transient hotspots and thermal gradients are adequately resolved. Features smaller than this diffusion length would naturally blur due to heat spreading, so a finer spatial resolution would not yield significantly more information. For larger dies or package-scale modeling, we plan to explore hierarchical training strategies with adaptive spatial resolutions.

\textit{COMSOL} captures both steady-state and transient heat flow dynamics, as illustrated in Fig.~\ref{fig:data}(b), by providing time-resolved temperature gradients, heat flux distributions, and potential hotspot identification across multiple time steps. The dataset covers a temperature range of approximately 25$^{\circ}$C to 55$^{\circ}$C, targeting nominal and moderately stressed operating regimes typically observed during early design phases such as placement and floorplanning. 
While thermal reliability assessments often extend beyond 85$^{\circ}$C for post-layout signoff, our focus here is on early-stage hotspot detection and thermal pattern learning. To ensure generalization beyond this nominal range, we additionally performed cross-validation on an extended small dataset spanning 25–95$^{\circ}$C as discussed in Section~\ref{cv_result}. The dataset can be found in the link: \href{https://purdue0-my.sharepoint.com/:f:/g/personal/chand133_purdue_edu/IgAB4G-8tmVgQ5PNUFUJt_IVASPrY6_te7pSm8-dmKiQgn4}{\textbf{\textcolor{blue}{2D-ThermAl Dataset}}}.

To ensure high-fidelity thermal simulations, it is crucial to define boundary conditions and material properties with precision. In our simulations, the die and substrate are assigned thermophysical parameters—such as thermal conductivity, density, and specific heat capacity—based on experimentally measured or literature-sourced values. Rigorous boundary condition settings, including insulated walls (to restrict external heat loss) and appropriate thermal loading, ensure accurate modeling of heat retention and dissipation under operational stresses. By integrating these thermally rich datasets with electrical insights from \textit{CADENCE}, we create a multidimensional dataset that captures the intricate interaction between electrical and thermal characteristics, forming a robust foundation for training our generative AI model.

\section{Methodology} \label{methods}
\begin{figure*}[!ht]
    \centering
    \includegraphics[width=\linewidth]{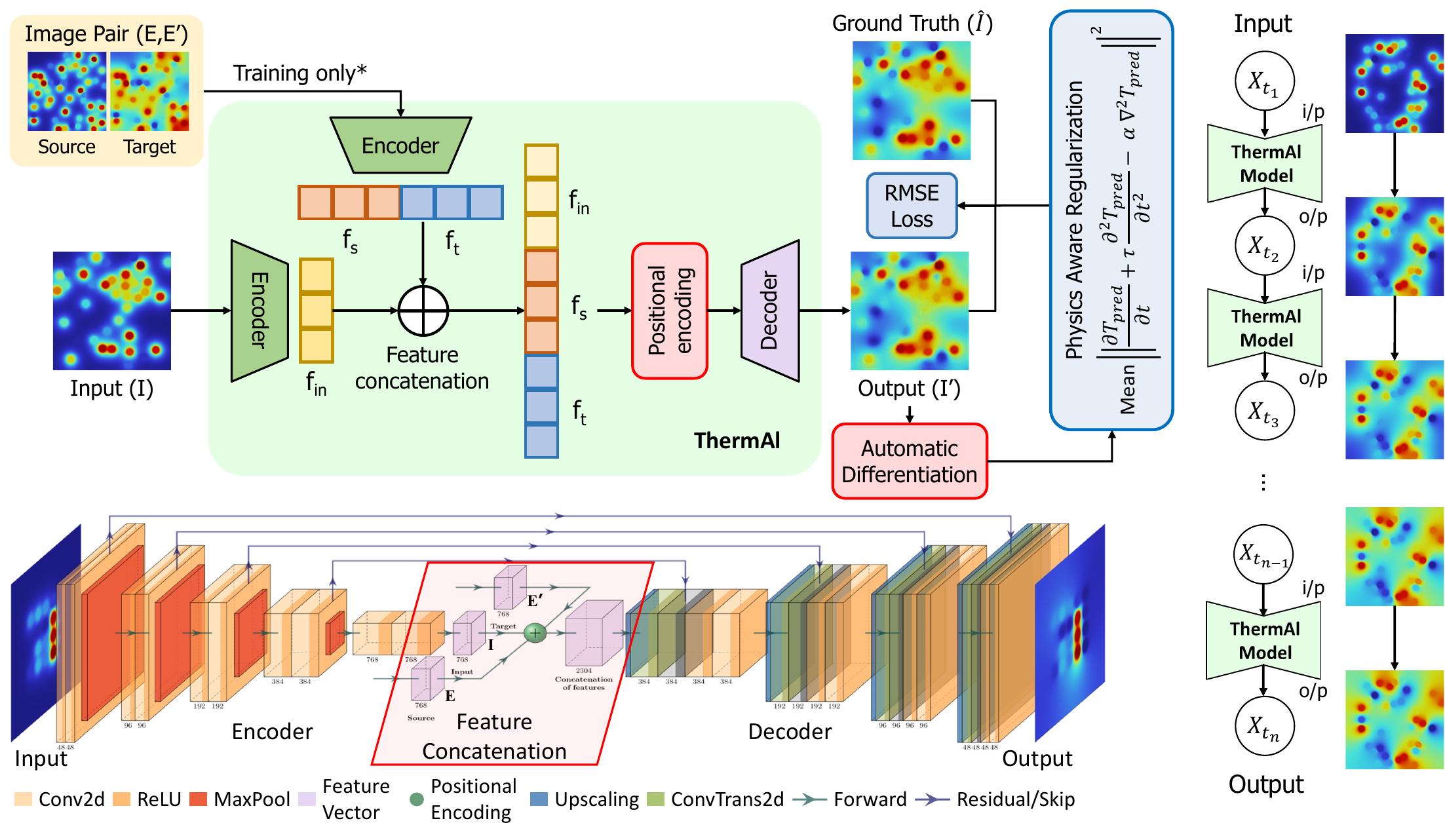}
    \vspace{-1em}
    \caption{Overview of the ThermAl framework: A novel hybrid U-Net network combining CNN and positional embedding for feature extraction and processing. The network accepts a pair of source-target images $\{E: \text{source}, E^\prime: \text{target}\}$, along with a input query image, $\{I\}$, and processes them through separate UNet-based encoders to extract multi-scale features. The resulting feature maps are concatenated to form $f_c$, which is then augmented with positional embeddings $f_{pos}$ and fed into a decoder to generate a dense prediction. The framework is trained using a composite loss function that combines the RMSE loss with a physics-aware regularizer, ensuring that predictions adhere to the underlying heat conduction dynamics. The flow-diagram illustrates how the model predicts transient heat dissipation and thermal flow over various time intervals, capturing the temporal evolution of thermal behavior.}
    \vspace{-0.5em}
    \label{fig:therm_model}
\end{figure*}

We reframe thermal analysis as an image-to-image translation task, extending the classic U-Net framework~\cite{ronneberger2015u} to address heatflow-specific challenges. Although the original U-Net excels at segmentation, it lacks certain mechanisms required for the dense spatial transformations intrinsic to thermal analysis. Our objective is to predict an output image $I^\prime$ from a query image $I$, given a pair of sample images ${E, E^\prime}$. We adopt feature-level concatenation and positional embedding (illustrated in Fig.~\ref{fig:therm_model}) to represent the intricate spatial relationships governing heat diffusion across different parts of the circuit. These enhancements allow the network to embed higher level semantic context, enabling it to predict both steady-state and transient thermal states in complex workloads. The following subsections describes each component of our model: input data encoding, feature concatenation with positional embedding, image reconstruction, and our physics-informed loss function.

\subsection{Input Data and Encoder} 
The model input comprises a source–target image pair ${E, E^\prime}$ and a query image $I$, each representing 2D thermal heatmaps of size $H \times W$, where pixel intensity encodes temperature gradients. Here, $I$ corresponds to the current thermal state provided as model input, while $E$ and $E^\prime$ represent reference thermal maps from current and desired time steps, respectively, sampled under different workloads. During training, the reference pair ${E, E^\prime}$ serves as contextual guidance, enabling the network to learn temporal transitions and spatial correspondences across thermal states. Each image is processed by an independent encoder to extract high-level feature representations: $
f_s = \mathrm{Enc}(E), \;
f_t = \mathrm{Enc}(E^\prime), \;
f_{in} = \mathrm{Enc}(I),
$
where $f_s$, $f_t$, and $f_{\text{in}} \in \mathbb{R}^{H \times W \times C}$ denote the encoded feature maps for the source, target, and query images, respectively. Each encoder employs dilated convolutions to capture both multi-scale spatial structures and localized temperature variations. Progressive downsampling is applied to reduce spatial resolution while preserving essential information about temperature gradients, yielding compact, high-level feature embeddings.

\subsection{Feature Concatenation and Positional Encoding}
A key architectural design choice is the use of feature-space concatenation rather than pixel-level concatenation to model spatial dependencies. The encoded query feature map $f_{in}$ is concatenated channel-wise with the source $f_s$ and target $f_t$ feature maps to form a unified content representation: $(f_c = \mathrm{Concat}(f_{in}, f_s, f_t))$, where $f_c \in \mathbb{R}^{H \times W \times 3C}$ aggregates spatial context from the reference maps ${E, E^\prime}$, enriching the query representation and improving transformation inference. To enhance geometric consistency, a positional encoding function $f_{pos}(x, y)$ is introduced to embed explicit spatial coordinates:
\begin{align}\label{eq:pos}
f_{\text{pos}}(x, y) = \Bigl[
& \sin\bigl(\tfrac{x}{10000^{2i/d}}\bigr), 
\cos\bigl(\tfrac{x}{10000^{2i/d}}\bigr), \nonumber\\
& \sin\bigl(\tfrac{y}{10000^{2i/d}}\bigr), 
\cos\bigl(\tfrac{y}{10000^{2i/d}}\bigr)
\Bigr]
\end{align}
where $(x, y)$ are pixel coordinates, $i$ denotes the dimension index, and $d$ is the total encoding dimensionality. The positional encoding is concatenated with the content feature map to produce the merged feature tensor $(f_m = \mathrm{Concat}(f_c, f_{pos}))$ which preserves both absolute coordinate information and learned feature correlations. By embedding positional cues directly into the feature space, the network achieves improved spatial alignment and more accurate transformation prediction.

\subsection{Decoder and Image Reconstruction} 
The concatenated feature map $f_c$, enriched with positional encodings, is passed through a decoder module that reconstructs the thermal map at the original resolution. The decoder employs transposed convolutions with a stride of 2 and a $3 \times 3$ kernel at each upsampling stage to progressively restore spatial dimensions. Skip connections from corresponding encoder layers are integrated to preserve fine-grained spatial details, ensuring that localized temperature variations are accurately recovered. To further enhance reconstruction quality and suppress checkerboard artifacts, the upsampled outputs are optionally refined using bilinear interpolation and anti-aliasing filters. The final output of the decoder, denoted as $I'$, is spatially aligned with the query input $I$ and trained to approximate the ground truth map $\hat{I}$, enabling precise prediction of heat distribution under complex thermal flow conditions.

\subsection{Loss Function with Physics-Aware Regularization}
Our training objective combines pixel-level accuracy with physics-based consistency to ensure both visual fidelity and thermodynamic realism. The primary loss component is a pixel-wise Root Mean Squared Error (RMSE) that measures discrepancies between the predicted thermal map $I'$ and the ground truth $\hat{I}$:
\begin{equation}\label{eq:rmse}
\mathcal{L}_{\mathrm{RMSE}} = 
\sqrt{\frac{1}{H \times W} 
\sum_{i=1}^{H} \sum_{j=1}^{W} 
\bigl(I'(i, j) - \hat{I}(i, j)\bigr)^2}.
\end{equation}

Minimizing $\mathcal{L}_{\mathrm{RMSE}}$ (or equivalently, $\ell_2$ loss) encourages the network to reproduce average temperature patterns and achieve high pixel-level fidelity. However, such data-driven loss functions do not enforce the underlying heat conduction physics, particularly in transient regimes, nor guarantee that the predicted temperature fields obey causality or finite propagation speeds. To address this limitation, we introduce a \emph{physics-aware} regularization term inspired from a simplified Boltzmann Transport Equation (BTE) framework. Rather than invoking the full BTE, which requires strong assumptions about phonon distributions, we employ the Cattaneo–Vernotte (CV) form of the hyperbolic heat equation under the relaxation-time approximation, assuming near-equilibrium phonon transport. This formulation better captures finite propagation speed and thermal relaxation effects than classical Fourier diffusion, which assumes instantaneous heat conduction. The corresponding residual loss measures the deviation of the predicted temperature field $T_{\mathrm{pred}}$ from this conduction model: 
\begin{equation}\label{eq:bte}
\mathcal{L}_{\mathrm{physics}} = 
\mathrm{Mean}\Big\lVert 
\tfrac{\partial T_{\mathrm{pred}}}{\partial t} 
+ \tau\, \tfrac{\partial^2 T_{\mathrm{pred}}}{\partial t^2} 
- \alpha\, \nabla^2 T_{\mathrm{pred}}
\Big\rVert^2,
\end{equation}
where $\tau$ denotes the characteristic relaxation time (approximately $10^{-13}$~s for Si) and $\alpha$ represents the thermal diffusivity (on the order of $10^{-4}$~m$^2$/s). This residual explicitly penalizes physically inconsistent heat dynamics, ensuring the model adheres to non-Fourier, finite-speed conduction behavior. 
The final composite objective balances data fidelity and physics coherence:
\begin{equation}
\mathcal{L} = \mathcal{L}_{\mathrm{RMSE}} + \lambda\, \mathcal{L}_{\mathrm{physics}},
\end{equation}
where $\lambda$ is a weighting factor controlling the trade-off between reconstruction accuracy and physical plausibility.
As shown in our ablation study (Table~\ref{ablation_physics}), removing the physics-aware term leads to higher prediction errors, particularly in high-gradient or fast-transient regions, confirming that it enhances both boundary accuracy and physical consistency beyond raw data fitting.

\section{Experimental Setup} \label{experiment}
This section outlines the experimental setup used to evaluate the proposed thermal prediction framework. We describe the training configuration and implementation details, followed by an overview of the evaluation metrics and analysis methods. All experiments are designed to ensure reproducibility and to reflect realistic thermal behavior in integrated circuit environments.

\subsection{Training Procedure}
We train our network in a supervised learning setting, pairing source and target thermal images. The complete model is implemented in Python~3.7 using the PyTorch framework (version~2.3.1)~\cite{pytorch}, a widely adopted open-source library for machine learning. Training is conducted for 100~epochs on a Linux server equipped with an AMD~EPYC~7502~32-Core Processor and four Nvidia~A40 GPUs, using a batch size of~50. The Adam optimizer is employed with an initial learning rate of~$10^{-6}$ and momentum of~$0.999$, ensuring stable convergence. Our dataset comprises a large set of synthetic 2D image pairs tailored for robust image matching, with 75\% used for training and 25\% reserved for validation. As illustrated in Fig.~\ref{fig:therm_model}, the transient evolution of thermal distributions is tracked from the initial condition to the final steady state. Each training pair corresponds to consecutive frames $(t,\,t+\Delta t)$ from the same transient simulation, where $\Delta t \approx 1~ms$. During training, the model’s predicted output at time $(t+\Delta t)$ is compared against the ground truth frame (teacher signal) to compute the supervised one-step prediction loss. We do not unroll predictions for multi-step training. During inference, the model operates in a free-running rollout mode, where its own output is recursively used as input for subsequent steps without ground truth injection. To mitigate overfitting and enhance generalization, we apply regularization strategies such as data augmentation, early stopping, and dropout. Specifically, we use a dropout rate of~0.25, early stopping with a patience of~10~epochs, random Gaussian noise ($\sigma = 0.01$) for augmentation, and a learning rate decay factor of~0.9 every~20~epochs. These strategies collectively prevent over-parameterization, ensuring the model remains robust to unseen circuit configurations and diverse thermal conditions.

\vspace{-0.5em}
\subsection{Evaluation Metrics} \label{eval}
To evaluate the performance of our model in dense image matching, we utilize three primary metrics: \emph{Root Mean Squared Error} (RMSE), \emph{Normalized Pixel Difference} (NPD), and the \emph{Structural Similarity Index} (SSIM). These metrics collectively offer a comprehensive view of how precisely the model reconstructs images and captures their perceptual qualities.

\vspace{0.5em}
\subsubsection{RMSE (Root Mean Squared Error)}
The RMSE quantifies the square root of the average squared differences between the pixel intensities of the predicted image $I'$ and the GT image $\hat{I}$. By minimizing RMSE as defined in Eq.~\eqref{eq:rmse}, the model is guided to match the target at a pixel-by-pixel level.

\vspace{1em}
\subsubsection{NPD (Normalized Pixel Difference)}
NPD evaluates the average absolute deviation between $I'$ and $\hat{I}$, normalized by the maximum pixel value $\text{MAX}_I$. It provides an intuitive measure of relative error, bounded between 0 and 1; values closer to 0 suggest more accurate predictions:
\begin{equation} \label{eq:npd}
    \text{NPD} 
    = 
    \frac{1}{H \times W} 
    \sum_{i=1}^{H} 
    \sum_{j=1}^{W} 
    \frac{
        \bigl|\hat{I}(i, j) - I^\prime(i, j)\bigr|
    }{
        \text{MAX}_I
    }
\end{equation}

\vspace{0.5em}
\subsubsection{SSIM (Structural Similarity Index)}
The SSIM evaluates the perceptual similarity between the predicted and ground truth images by jointly considering \emph{structural information}, \emph{luminance}, and \emph{contrast} in images.
\begin{equation}\label{eq:ssim}
    \text{SSIM}(I^\prime, \hat{I}) 
    = 
    \frac{
        (2\mu_{I^\prime}\,\mu_{\hat{I}} + C_1)\,
        (2\,\sigma_{I^\prime, \hat{I}} + C_2)
    }{
        (\mu_{I^\prime}^2 + \mu_{\hat{I}}^2 + C_1)\,
        (\sigma_{I^\prime}^2 + \sigma_{\hat{I}}^2 + C_2)
    }
\end{equation}
where $\mu_{I^\prime}$ and $\mu_{\hat{I}}$ denote the local means of the predicted and GT images, $\sigma_{I^\prime}^2$ and $\sigma_{\hat{I}}^2$ are their respective local variances, and $\sigma_{I^\prime,\hat{I}}$ represents the local covariance. The constants $C_1$ and $C_2$ help stabilize the division. SSIM values range from $-1$ to $1$, with $1$ signifying perfect structural match. This metric is particularly adept at detecting perceptual differences due to variations in overall image structure.

\section{Results} \label{results}
\subsection{Thermal Map Estimation Accuracy}
Once \textit{ThermAl} is fully trained, it can process thermal maps from any time point as input and generate realistic full-chip temperature distributions, effectively predicting pixel-level variation across the chip. We evaluate the model's performance using the metrics outlined in Section~\ref{eval}, assessing both local and global deviations between generated thermal maps and ground truth (GT) data.

\begin{table}[!ht]
\centering
\caption{Performance metrics across different Transient stages}
\label{tab:result}
\resizebox{\linewidth}{!}{%
\begin{tabular}{|c|cc|c|c|c|}
\hline
\multicolumn{1}{|l|}{} & \multicolumn{2}{c|}{\textbf{Time/Epoch}} & \multicolumn{1}{c|}{\multirow{2}{*}{\textbf{NPD} $\downarrow$}} & \multicolumn{1}{c|}{\multirow{2}{*}{\textbf{RMSE} ($^\circ$\textbf{C}) $\downarrow$}} & \multicolumn{1}{c|}{\multirow{2}{*}{\textbf{SSIM} $\uparrow$}} \\ \cline{2-3}
\multicolumn{1}{|l|}{} & \textbf{Start} & \textbf{End} & \multicolumn{1}{c|}{} & \multicolumn{1}{c|}{} & \multicolumn{1}{c|}{} \\ \hline
\multirow{5}{*}{Intermediate} & 1 & 5 & 0.27 & 0.69 & 0.98 \\
 & 5 & 10 & 0.29 & 0.74 & 0.96 \\
 & 10 & 20 & 0.29 & 0.68 & 0.97 \\
 & 20 & 50 & 0.34 & 0.71 & 0.97 \\
 & 50 & 100 & 0.32 & 0.73 & 0.98 \\ 
\rowcolor[gray]{0.9}\multicolumn{1}{|l|}{Average} & 1 & 100 & 0.32 & 0.71 $\pm$ 0.26 & 0.97 $\pm$ 0.01 \\ \hline
\end{tabular}%
}
\end{table}

Thermal maps are represented as $256\times 256$ pixel image, where each pixel corresponds to a discrete temperature reading. Over the test set, we observe an average RMSE of \textbf{0.71$^\circ$C} (std. dev. 0.26$^\circ$C), as summarized in Table~\ref{tab:result}. The overall temperature range in these samples spans from 25$^\circ$C to 55$^\circ$C, which aligns with standard operating ranges for integrated circuitsd uring early design stages such as placement and floorplanning. Relative to this 30$^\circ$C variation window, ThermAl achieves an average full-scale estimation error of only \textbf{1.4\%} (std. dev. 0.25\%). These consistently low errors across diverse configurations highlight the model’s robust generalization, suggesting it is well-suited for real-world thermal monitoring tasks. 

\begin{figure}[!ht]
    \centering
    \includegraphics[width=\linewidth]{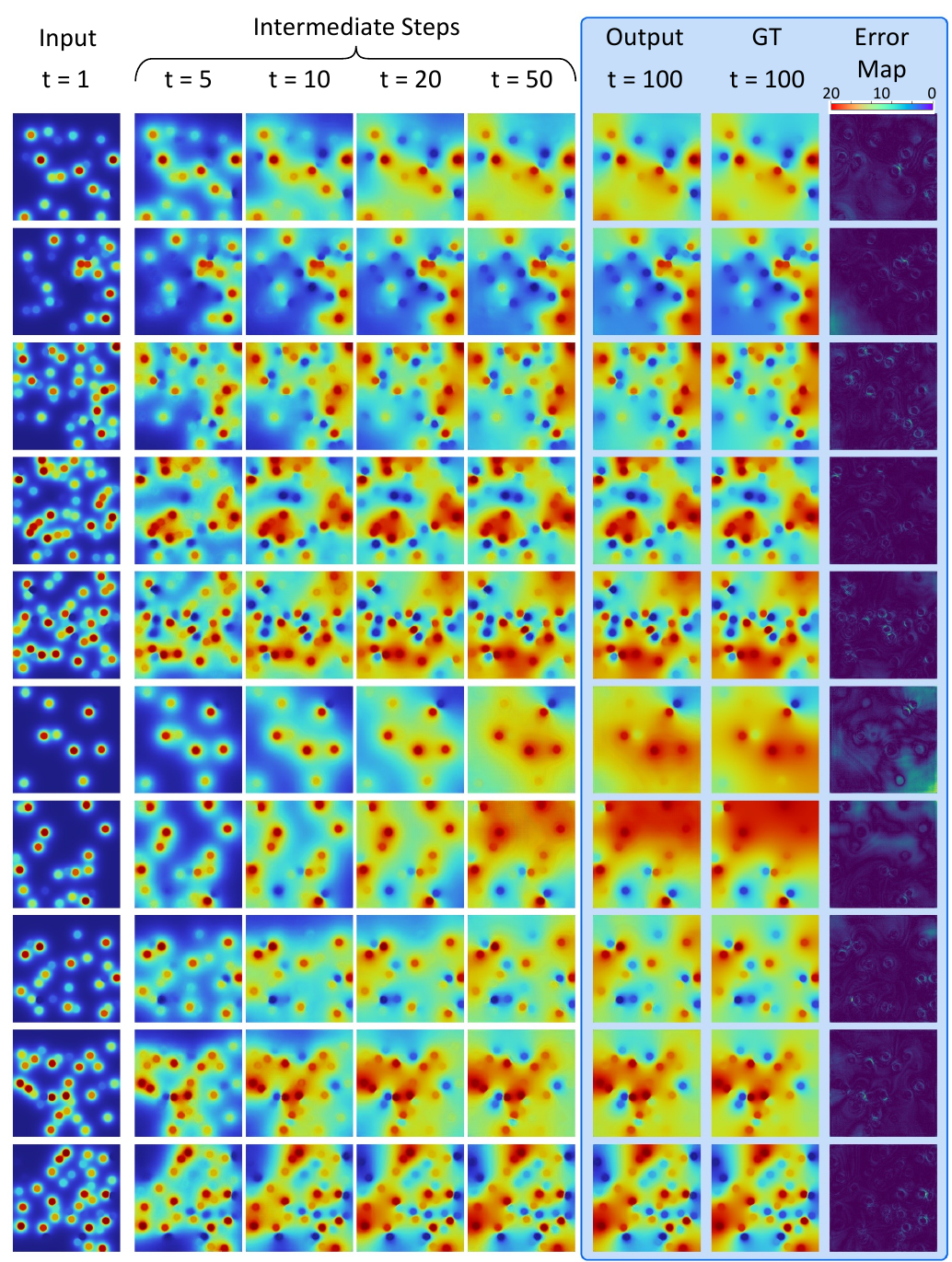}
    \caption{Qualitative results of ThermAl: The image demonstrates the generation of thermal maps from an initial sample (leftmost column), showing the progression of heat distribution over time (intermediate output columns). These steps capture the dynamic heat flow over time steps $t = 1, 5, 10, 20, 50$, with the final predicted output map at $t = 100$ compared against the ground truth (GT) in the rightmost columns. The error maps highlight the pixel-wise differences between the predicted and true simulated thermal maps.}
    \vspace{-1.5em}
    \label{fig:result1}
\end{figure}

Figure~\ref{fig:result1} illustrates how \textit{ThermAl} estimates a steady-state thermal map by progressing through multiple transient steps. Each intermediate stage captures dynamic heat flow interactions across the chip surface, reflecting evolving diffusion patterns dictated by the FEM boundary conditions. The model accurately reconstructs steady-state maps while maintaining low error rates, even in regions characterized by high heat flux and steep thermal gradients. In most regions, prediction errors remain within 0.61$^\circ$C and rarely exceed 1$^\circ$C, demonstrating the precision of the model in modeling uniform and non-uniform heat distributions.

In the present study, validation is predominantly conducted using high-fidelity synthetic datasets. It allows exhaustive coverage of thermal conditions and circuit configurations, from simple logic gates to complex workloads, at high resolutions and accuracy levels that are often prohibitively costly or practically unattainable with experimental thermal imaging techniques, such as infrared (IR) thermography, due to resolution limitations and calibration complexities. However, validation against real-world experimental data remains an essential step for practical applicability.

\subsection{Cross-Validation on Extended Thermal Range} \label{cv_result}
To further validate \textit{ThermAl}’s robustness under broader operating conditions, we performed cross-validation experiments on an independent dataset comprising 500 thermal maps generated from diverse input activity patterns across 100 circuit instances. Unlike the main evaluation set, this dataset spans a wider thermal range of 25$^\circ$C to 95$^\circ$C, designed to emulate high-power workloads and stress conditions characterized by stronger spatial gradients. Despite this 70$^\circ$C full-scale variation, \textit{ThermAl} sustains high predictive fidelity, achieving an average RMSE of \textbf{1.5$^\circ$C}, corresponding to less than \textbf{2.2\%} full-scale error. These results confirm that the model generalizes effectively to unseen operating regimes while preserving fine-grained, pixel-level accuracy.

\begin{figure}[!ht]
\centering
\includegraphics[width=\linewidth]{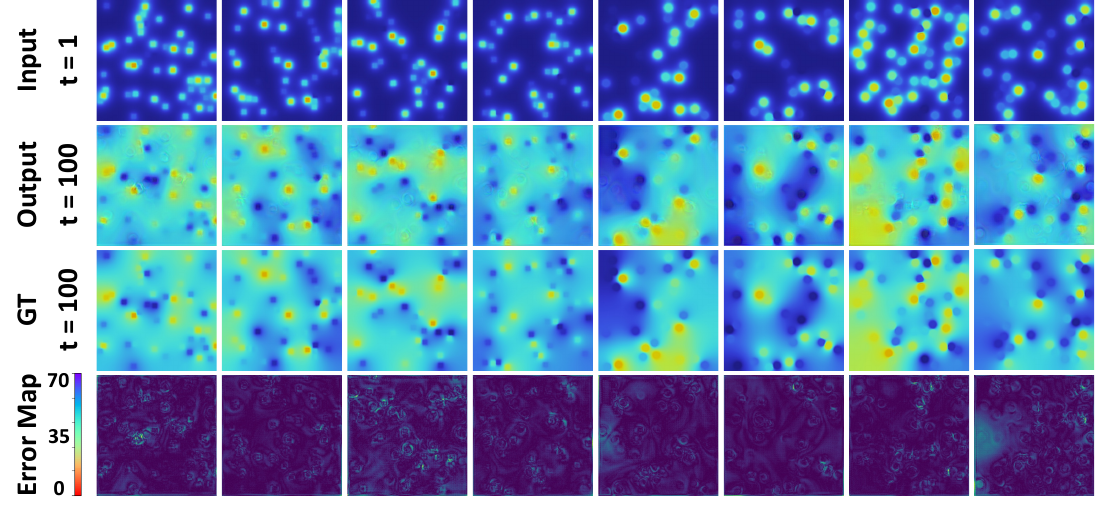}
\caption{Qualitative results from the extended-range cross-validation dataset. Each row displays: (a) the input at time $t = 0$, (b) the ground-truth thermal map, (c) the \textit{ThermAl}-predicted map, and (d) the absolute error map. The model shows strong spatial consistency and maintains low residual errors even across a wide temperature range (25--95$^\circ$C).}
\vspace{-1em}
\label{fig:cv_error_maps}
\end{figure}

To complement the quantitative analysis, Figure~\ref{fig:cv_error_maps} presents qualitative error maps illustrating prediction fidelity across representative samples. Each row includes: (a) the input at time $t = 0$, (b) the ground-truth thermal map, (c) the corresponding \textit{ThermAl} prediction, and (d) the pixel-level absolute error distribution. The results show that errors remain spatially uniform across most of the chip, with slightly elevated deviations concentrated near localized hotspots where transient gradients are steepest. Even in these high-variation regions, the deviation rarely exceeds 2$^\circ$C, indicating that \textit{ThermAl} effectively captures both uniform and non-uniform heat propagation patterns.

These findings directly address concerns regarding generalization and practical deployment. The model’s ability to maintain low error over a substantially wider temperature span demonstrates that it is not overfitted to a narrow synthetic domain. Instead, it adapts effectively to more realistic and diverse conditions. This cross-validation experiment strengthens confidence in \textit{ThermAl}’s applicability for early-stage thermal modeling and real-time chip monitoring, paving the way for seamless integration with experimental validation.

\begin{table*}[ht!]
    \centering
    \caption{Comparison with Representative State-of-the-Art (SOTA) Thermal Modeling Frameworks}
    \label{tab:sota_comparison}
    \resizebox{\linewidth}{!}{
    \begin{tabular}{|l|c|c|c|c|c|c|}
    \hline
    \textbf{Approach} & \textbf{Methodology} & \textbf{Latency} & \textbf{RMSE (}$^\circ$\textbf{C)} $\downarrow$ & \textbf{Grid / Model Size} & \textbf{Speedup vs FEM} & \textbf{Key Limitations} \\ \hline
    \multicolumn{7}{|c|}{\textbf{Solver- and Compact-Model-Based Methods}} \\ \hline
    COMSOL~\cite{comsol} & FEM Solver & $\geq$3--4 s & Reference & Any mesh & 1$\times$ & High resource usage; Non-real-time \\
    HotSpot~\cite{hotspot} & Thermal RC Network & 25.8 ms & 2--4.2$^\circ$C & $\sim$160$\times$160 & $\sim$150$\times$ & Manual calibration; coarse grid \\
    Power Blurring~\cite{powerblurring} & Green's Function Convolution & $\sim$100 ms & $\leq$2\% & 192$\times$192 & $\sim$40$\times$ & Static kernel; limited dynamic flexibility \\
    PACT~\cite{PACT} & Compact Thermal Solver & $\sim$200 ms & $\leq$3.3\% & Up to 512$\times$512 & $\sim$20$\times$ & Parallel solver; setup overhead \\
    3D-ICE~\cite{3DICE} & Compact 3D Solver & $\sim$50 ms & $\leq$3.4$^\circ$C & Full 3D Stack & $\sim$975$\times$ & Requires detailed package model \\
    MTA~\cite{MTA} & Numerical / Compact Hybrid & 30--40 s (steady) 
    & $\sim$2$^\circ$C & Multi-layer 3D & 100$\times$--1000$\times$ & Heavy pre-processing; slower for fine steps \\ \hline
    \multicolumn{7}{|c|}{\textbf{Learning-Based Methods}} \\ \hline
    LSTM+DCT~\cite{sadiqbatcha2020machine} & ML Regression & 19 ms & $\leq$2.26\% & 128$\times$128 & $\sim$200$\times$ & Requires historical data \\
    ThermGAN~\cite{jin2020full} & GAN (Generative Model) & 7.5 ms & 0.47$^\circ$C & 128$\times$128 & $\sim$300$\times$ & Needs IR camera / paired data \\
    DeepOHeat~\cite{deepoheat} & Neural Operator Learning & $\sim$5 ms & 0.1--0.5$^\circ$C & 3D domain & 10$^3$--10$^5\times$ & Requires 3D dataset; large training cost \\
    \rowcolor[gray]{0.9}
    \textbf{ThermAl (Ours)} & \textbf{NN + Physics Regularizer} & \textbf{$\leq$10 ms} & \textbf{0.71$^\circ$C (1.4\%)} & \textbf{256$\times$256} & \textbf{$\sim$200$\times$} & Limited boundary modeling; 2D only \\ \hline
    \end{tabular}
    }
    \vspace{0.5em}
    {\\ \scriptsize \textit{* Reported metrics are approximate and depend on hardware, boundary conditions, and dataset configurations. FEM runtimes (COMSOL) used as baseline for relative speedups.}}
    \vspace{-1em}
\end{table*}

\vspace{-0.25em}
\subsection{Comparison to Commercial FEM Tools}
While training \textit{ThermAl} requires substantial computational resources (approximately 18~hours to reach convergence), inference operates in near real time. Once trained, the full simulation—from initial input to steady-state prediction—completes within 7–10~milliseconds per sample, corresponding to an inference rate exceeding 100~thermal maps per second. In contrast, commercial FEM solvers such as \textit{COMSOL}~\cite{comsol} provide high physical fidelity but at the cost of long runtimes (3–4~seconds per map) under comparable hardware and workload conditions, limiting their practicality for iterative design exploration. Although optimized numerical solvers offer moderate runtime improvements, as shown in Table~\ref{tab:sota_comparison}, the millisecond-level inference capability of \textit{ThermAl} delivers an effective speedup of approximately $200\times$, offering a significant advantage for large-scale design workflows that require rapid thermal feedback and iterative optimization.

\subsection{Comparison with State-of-the-Art Models}
Beyond commercial FEM solvers, we benchmark \textit{ThermAl} against several contemporary machine learning (ML) and semi-analytical approaches to evaluate its efficiency, scalability, and predictive accuracy. Table~\ref{tab:sota_comparison} summarizes this comparison in terms of latency, accuracy, and modeling assumptions across representative techniques. Overall, \textit{ThermAl} achieves a favorable balance between computational speed, physical interpretability, and ease of deployment. 

Compact thermal modeling frameworks such as \textit{HotSpot}~\cite{hotspot}, \textit{PACT}~\cite{PACT}, and \textit{3D-ICE}~\cite{3DICE} offer faster runtimes but rely heavily on manual calibration and explicit package modeling. ML-based thermal estimators further reduce latency but often depend on specialized infrared (IR) imaging systems or dense sensor arrays~\cite{jin2020full}, exhibit limited scalability for 3D heat-flow analysis~\cite{powerblurring}, or require extensive historical sensor data~\cite{sadiqbatcha2020machine}. Operator-learning frameworks such as \textit{DeepOHeat}~\cite{deepoheat} achieve extreme acceleration—up to $10^{3}$–$10^{5}\times$ faster than FEM—with minimal steady-state error; however, they face challenges in scalability, model complexity, and generalization when boundary conditions or material parameters change.

Within this landscape, \textit{ThermAl} establishes an effective middle ground between high-fidelity physics-based simulators and data-intensive operator-learning models. It achieves sub-degree accuracy (RMSE~$\approx~0.71~^\circ$C) with millisecond-level inference latency, eliminating the need for boundary reconfiguration or retraining. These characteristics make it particularly suitable for early-stage placement and floorplanning tasks in thermal-aware electronic design automation (EDA). Although the current implementation abstracts heat-sink dynamics and assumes fixed boundary resistance, \textit{ThermAl} distinguishes itself through its balanced trade-off among speed, accuracy, and integration simplicity—offering a practical and scalable solution for rapid, layout-aware thermal estimation in modern design workflows.

\subsection{Extending Modeling to Complex Workloads}
\begin{figure}[!ht]
    \centering
    \includegraphics[width=\linewidth]{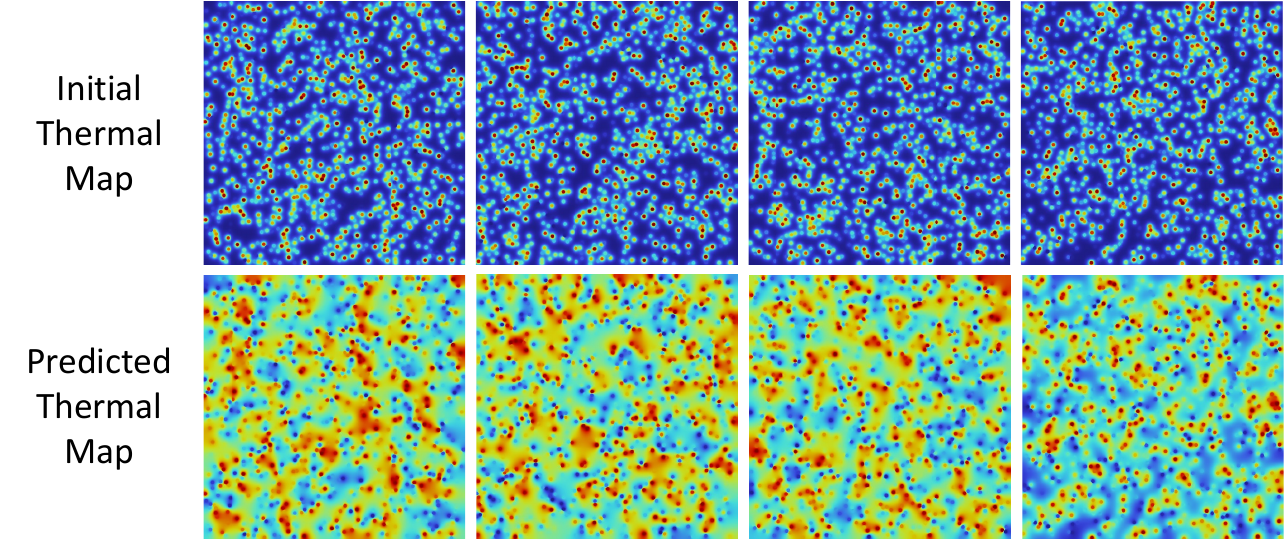}
    \caption{\textit{ThermAl} identifying hotspots across challenging workloads. Its scalability ensures detailed temperature monitoring even under high-complexity scenarios.}
    \vspace{-1em}
    \label{fig:result3}
\end{figure}

Finally, \textit{ThermAl} extends beyond circuit-level tasks to handle high-complexity workloads and higher spatial resolutions, as demonstrated in Figure~\ref{fig:result3}. In this setting, the model processes large-scale, heterogeneous circuit layouts at resolutions up to $512 \times 512$, enabling fine-grained prediction of temperature gradients. Despite the increased resolution and complexity, \textit{ThermAl} maintains strong predictive performance, achieving an average RMSE of only 1.08$^\circ$C. This capability supports proactive power-thermal management strategies and offers deeper insights into thermal behavior at the system level.

\section{Ablation Study} \label{ablation}
\begin{figure}[!ht]
    \centering
    \includegraphics[width=\linewidth]{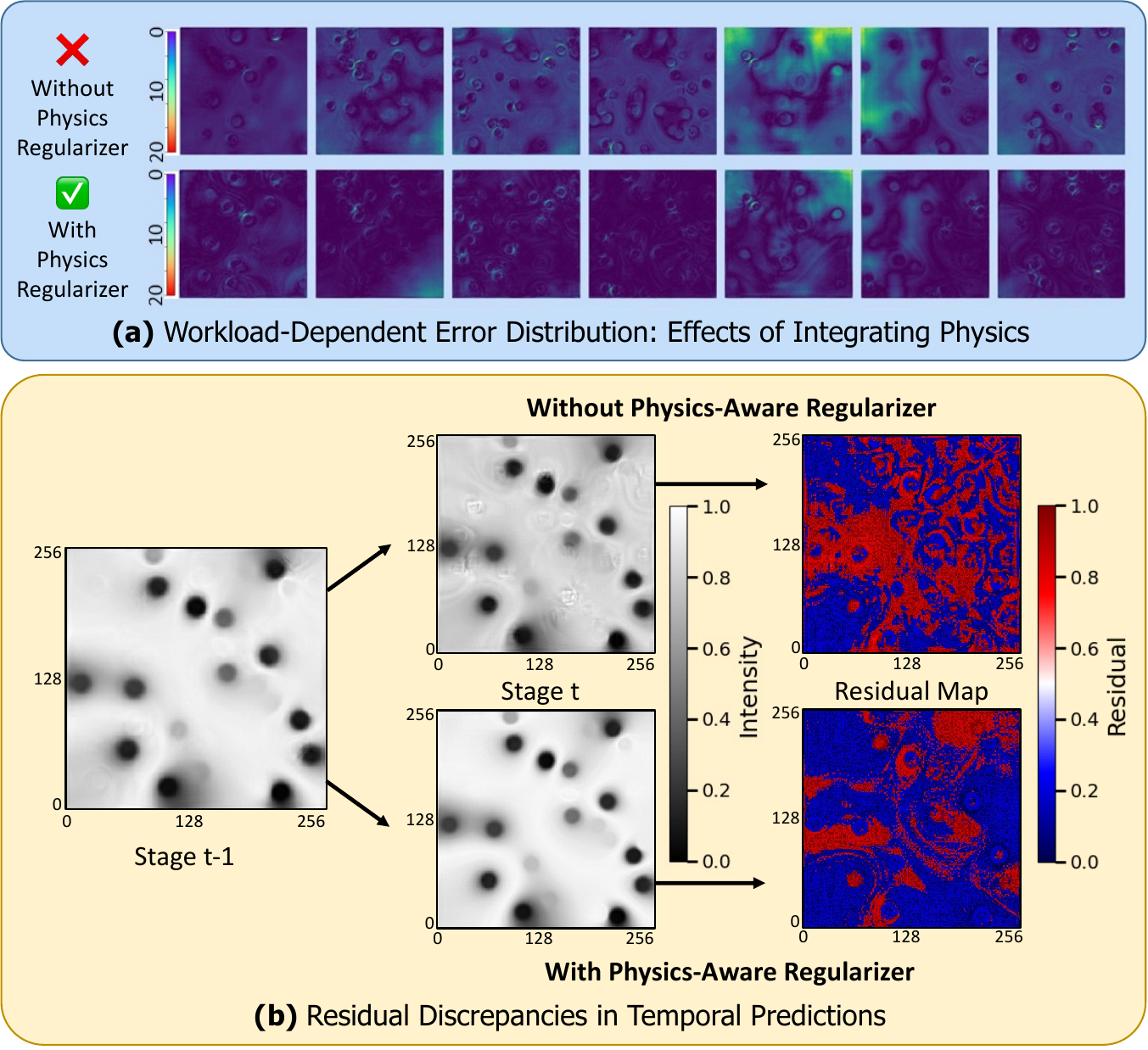}
    \caption{Visualizing the Impact of Physics-Aware Regularization. (a) Workload-Dependent Error Distribution: The top row displays error maps without the physics-based loss, revealing higher inaccuracies in complex hotspots; the bottom row demonstrates how physics constraints notably reduce these deviations. (b) Residual Discrepancies in Temporal Predictions: Transition stages $t-1$ to $t$ show that the physics term diminishes spurious residuals in high-flux regions, leading to more stable and physically consistent predictions.}
    \label{fig:physics_ablation}
    \vspace{-1em}
\end{figure}

\noindent Beyond our baseline RMSE-driven training procedure, we incorporate three major enhancements to improve thermal prediction fidelity:
\begin{enumerate}
    \item Physics-based regularizer.
    \item Source-target image pair.
    \item Feature-level concatenation.
\end{enumerate}
This section examines each improvement in detail, supported by both quantitative results (Tables~\ref{tab:exp_results} and~\ref{tab:ablation2}) and visual evidence (see Fig.~\ref{fig:physics_ablation}).

\vspace{-0.5em}
\subsection{Impact of Physics-Aware Learning} \label{ablation_physics}
Heat conduction in solids is fundamentally governed by partial differential equations (PDEs) that describe both spatial and temporal temperature evolution. To ensure that our model remains consistent with these physical principles, we incorporate a physics-informed regularization term based on a simplified form of the Boltzmann Transport Equation (BTE). Specifically, we employ the Cattaneo–Vernotte (CV) formulation of the hyperbolic heat equation under the relaxation-time approximation, which captures finite thermal propagation speeds and transient relaxation effects beyond classical Fourier diffusion. This additional loss term stabilizes training and improves prediction fidelity, particularly in regions with steep temperature gradients or rapid transients.

As illustrated in Fig.~\ref{fig:physics_ablation}(a), incorporating the physics-aware loss substantially reduces spurious artifacts and produces smoother, more physically consistent thermal maps. The residual error maps in Fig.~\ref{fig:physics_ablation}(b) further reveal how the inclusion of physical constraints mitigates localized deviations from expected heatflow dynamics. Quantitatively, as summarized in Table~\ref{tab:exp_results}, the physics-aware regularizer lowers the RMSE from 1.12~$^\circ$C to 0.76~$^\circ$C and the normalized pixel difference (NPD) from 0.42 to 0.32. These improvements are particularly pronounced near high-gradient regions, confirming that the added physical constraint enhances model robustness and boundary accuracy beyond raw data fitting.

While this BTE-inspired approach effectively models phonon transport at the microscale, its accuracy diminishes at sub-7~nm technology nodes, where quantum confinement, ballistic transport, and grain boundary scattering introduce non-local effects that classical PDE-based formulations cannot fully capture. Future extensions of \textit{ThermAl} may incorporate quantum-aware transport approximations or learned phonon interaction models to improve nanoscale thermal accuracy. Although the current physics-aware formulation slightly increases model size (up to 450~MB), the corresponding gains in physical realism and prediction stability justify this trade-off. Further enhancements such as adaptive mesh refinement, multi-scale learning, or higher-order regularization terms could provide additional robustness in capturing abrupt hotspot transitions and fine-grained thermal behavior.

\begin{table}[ht]
    \centering
    \vspace{-0.5em}
    \caption{Experimental Results: Impact of Source-Target Image Pair and Physics Regularizer}
    \label{tab:exp_results}
    \resizebox{\linewidth}{!}{
    \begin{tabular}{|cc|ccc|c|c|}
    \hline
    \textbf{Image} & \textbf{Physics} & \textbf{NPD} $\downarrow$ & \textbf{RMSE (}$^\circ$\textbf{C)}$\downarrow$ & \textbf{SSIM}$\uparrow$ & \textbf{Model Size} & \textbf{Params}\\ \hline
     &  & 0.42 & 1.12 $\pm$ 0.43 & 0.91 & 330 MB & 17M \\
    \checkmark &  & 0.36 & 0.84 $\pm$ 0.24 & 0.95 & 446 MB & 23M \\
     & \checkmark & 0.35 & 0.76 $\pm$ 0.18 & 0.96 & 333 MB & 17.4M \\
    \rowcolor[gray]{0.9}\checkmark & \checkmark & 0.32 & 0.71 $\pm$ 0.26 & 0.97 & 450 MB & 23.6M \\ \hline
    \end{tabular}
    }
    \vspace{-1em}
\end{table}

\subsection{Impact of condition-based learning via Source-Target Image Pairs}
Our second enhancement uses a source-target image pair to provide contextual references for how thermal patterns should evolve. By seeing a pair $\{E, E'\}$ in addition to the query $I$, the model learns to map initial and final states of the heat distribution. Table~\ref{tab:exp_results} shows that merely introducing the image pair lowers the RMSE from 1.12$^\circ$C to 0.84$^\circ$C and boosts the SSIM from 0.91 to 0.95. Even without the physics constraint, condition-based learning alone yields a notable improvement in capturing both local and global heat transfer processes.

\begin{table}[!ht]
\centering
\vspace{-0.5em}
\caption{Feature-level versus pixel-level concatenation comparison}
\label{tab:ablation2}
\begin{tabular}{|l|ccc|}
\hline
 & NPD $\downarrow$ & RMSE ($^\circ$C) $\downarrow$ & SSIM $\uparrow$ \\ \hline
Pixel-level & 0.45 & 1.18 $\pm$ 0.72  & 0.87 $\pm$ 0.02  \\
\rowcolor[gray]{0.9}Feature-level & 0.32 & 0.71 $\pm$ 0.26 & 0.97 $\pm$ 0.01 \\ \hline
\end{tabular}
\vspace{-1em}
\end{table}

\subsection{Effectiveness of Feature-Level Concatenation}
Feature-level concatenation enhances our model by merging feature maps from the query, source, and target images at a deeper level to capture complex spatial and contextual relationships necessary for thermal image transformation, shown in Table~\ref{tab:ablation2}. While pixel-level concatenation is effective for segmentation tasks, it lacks the depth required to interpret both global patterns and nuanced local details for accurate thermal predictions. By operating within this feature space, our model can synthesize diverse types of information, such as gradients, textures, and abstracted heat flow relationships, leading to an informed representation for subsequent thermal predictions.

\section{Future Works and Conclusion} \label{conclusion}
The proposed \textit{ThermAl} framework offers a \emph{fast}, physically \emph{consistent}, and data-driven approach to on-chip thermal analysis by integrating generative modeling with physics-informed regularization and a comprehensive library of 2D thermal simulations. Capable of achieving high predictive accuracy (mean error of 0.71~$^\circ$C at $256 \times 256$ resolution) with inference times under 10~ms—up to $200\times$ faster than conventional FEM solvers—\textit{ThermAl} enables rapid thermal feedback during early-stage IC design. This capability significantly reduces the risk of late-stage thermal bottlenecks and supports efficient exploration of power-performance trade-offs within electronic design automation (EDA) workflows. The model demonstrates strong generalization across a diverse set of synthetic layouts and transient heat scenarios, making it a practical candidate for integration into thermal-aware placement and floorplanning tools.

Despite its promising results, several directions remain for future improvement. Enhancing prediction accuracy around hotspot boundaries remains a challenge due to steep gradients and abrupt material transitions. Future work will explore adaptive mesh refinement, boundary-aware loss functions, and multi-scale architectures to better capture localized variations. Moreover, the current implementation is limited to 2D heat diffusion and does not account for vertical heat transfer across stacked dies or packaging layers. Extending the framework to 3D thermal modeling using volumetric FEM datasets and 3D convolutional or transformer-based architectures is a key next step. \textit{ThermAl} has been developed and validated primarily on 65~nm technology data; evaluating its adaptability to advanced nodes and novel floorplan styles will further strengthen its applicability. Transfer learning and domain adaptation strategies will be explored to facilitate cross-node generalization with minimal retraining effort. Additionally, experimental validation using real-world measurements—such as infrared thermography and on-die temperature sensors—will be critical for establishing physical robustness under realistic operating conditions. 
In summary, \textit{ThermAl} provides a scalable and physically grounded framework for fast, layout-aware thermal prediction in integrated circuits. Its ability to deliver accurate, real-time thermal estimation enables early detection of potential hotspots and supports efficient power-performance trade-off exploration during design. Additional cross-validation over an extended thermal range further confirmed \textit{ThermAl}’s robustness, maintaining low error and stable generalization even under high-power, high-gradient conditions representative of real-world workloads. Together, these results highlight \textit{ThermAl}’s potential as a foundational step toward real-time thermal monitoring and active thermal management in next-generation chip designs, where increasing power densities and performance demands require both precision and adaptability.

\section{Biography Section}
\begin{IEEEbiography}[{\includegraphics[width=1in,height=1.25in,clip,keepaspectratio]{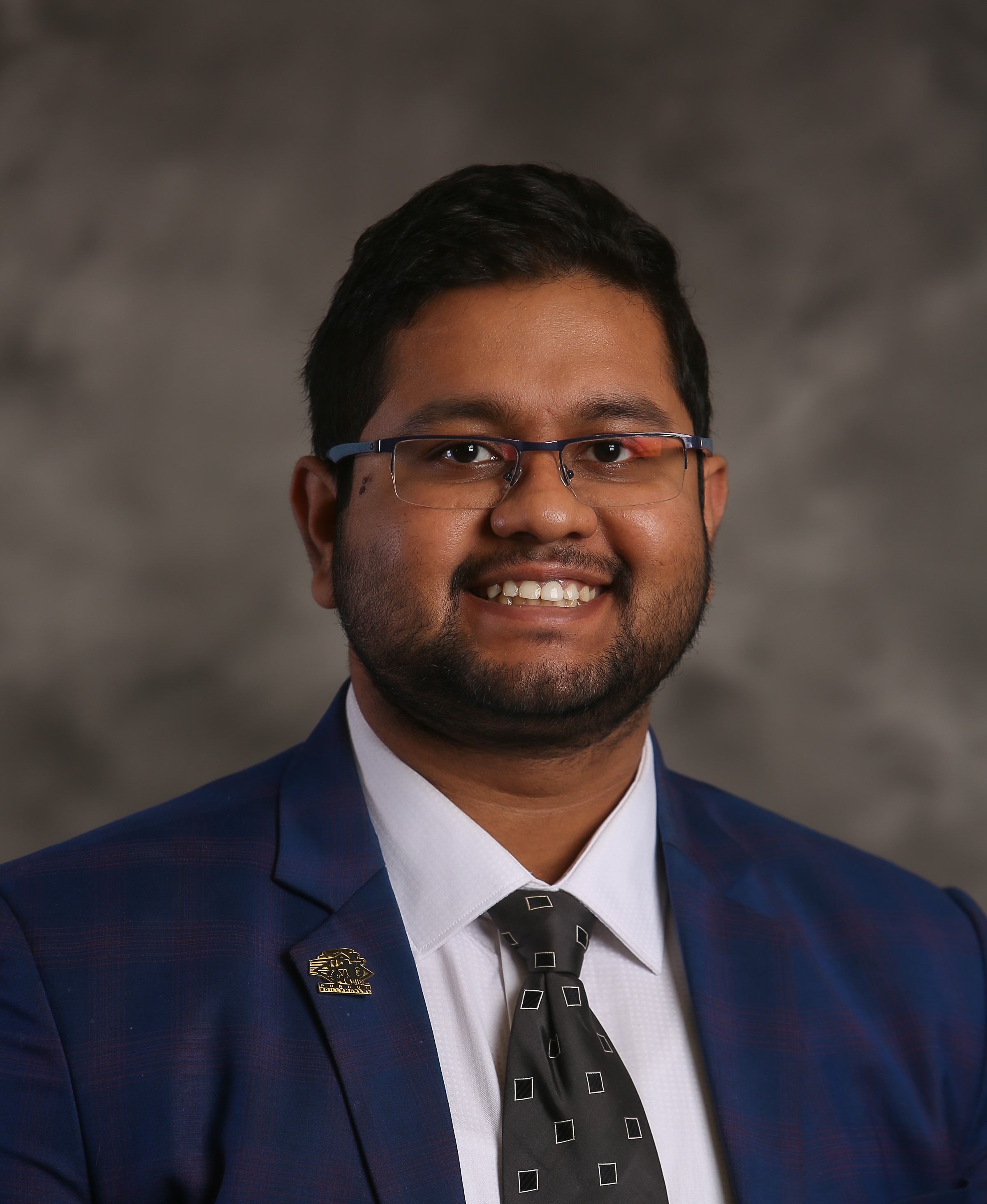}}]{Soumyadeep Chandra} received his B.Tech degree in Electrical and Telecommunication Engineering from Jadavpur University, Kolkata, India in 2020.

Currently, he is a 4th year Ph.D. student in the School of Electrical and Computer Engineering at Purdue University, West Lafayette, IN, under the mentorship of Prof. Kaushik Roy. His ongoing research focuses on areas such as computational biology, scene understanding, surgical workflow analysis, generative AI for hardware, and AI-driven thermal modeling. His work explores the intersection of machine learning and physical systems, leveraging data-driven approaches to predict and optimize thermal behavior in complex hardware architectures.
\end{IEEEbiography}

\begin{IEEEbiography}[{\includegraphics[width=1in,height=1.25in,clip,keepaspectratio]{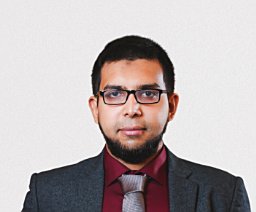}}]{Sayeed Shafayet Chowdhury} received his B.Sc. degree in Electrical \& Electronic Engineering from Bangladesh University of Engineering and Technology in 2016. He won the IEEE Signal Processing Cup (SP Cup) competition in 2015 and 2016, and also obtained 3rd position in the IEEE Video \& Image Processing Cup (VIP Cup) competition in 2017. 

He graduated a Ph.D. graduate student from the School of Electrical and Computer Engineering at Purdue University, West Lafayette, IN, under the guidance of Prof. Kaushik Roy. His current research interests include neuromorphic computing, specifically, developing energy-efficient algorithms for deep spiking neural networks (recognition, inference, analytics), visual reasoning, etc.
\end{IEEEbiography}

\begin{IEEEbiography}[{\includegraphics[width=1in,height=1.25in,clip,keepaspectratio]{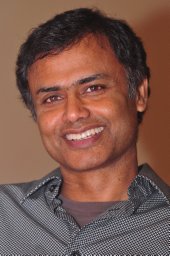}}]{Kaushik Roy}  (Fellow, IEEE) is the Edward G. Tiedemann, Jr., Distinguished Professor of Electrical and Computer Engineering at Purdue University. He received his BTech from the Indian Institute of Technology, Kharagpur, PhD from the University of Illinois at Urbana-Champaign in 1990, and joined the Semiconductor Process and Design Center of Texas Instruments, Dallas, where he worked for three years on FPGA architecture development and low-power circuit design. His current research focuses on cognitive algorithms, circuits, and architecture for energy-efficient neuromorphic computing/ machine learning, and neuro-mimetic devices. Kaushik has supervised more than 100 PhD dissertations and his students are well placed in universities and industry. He is the co-author of two books on Low Power CMOS VLSI Design (John Wiley \& McGraw Hill). 

Dr. Roy received the National Science Foundation Career Development Award in 1995, IBM faculty partnership award, ATT/Lucent Foundation award, 2005 SRC Technical Excellence Award, SRC Inventors Award, Purdue College of Engineering Research Excellence Award, Outstanding Mentor Award in 2021, Humboldt Research Award in 2010, 2010 IEEE Circuits and Systems Society Technical Achievement Award (Charles Desoer Award), IEEE TCVLSI Distinguished Research Award in 2021,  Distinguished Alumnus Award from Indian Institute of Technology (IIT), Kharagpur, Fulbright-Nehru Distinguished Chair, DoD Vannevar Bush Faculty Fellow (2014-2019), SRC Aristotle Award in 2015, Purdue Arden L. Bement Jr. Award in 2020, SRC Innovation Award in 2022, honorary doctorate from Aarhus University in 2023, and best paper awards at 1997 International Test Conference, IEEE 2000 International Symposium on Quality of IC Design, 2003 IEEE Latin American Test Workshop, 2003 IEEE Nano, 2004 IEEE International Conference on Computer Design, 2006 IEEE/ACM International Symposium on Low Power Electronics \& Design, 2005 and 2019 IEEE Circuits and system society Outstanding Young Author Award (Chris Kim, Abhronil Sengupta), 2006 IEEE Transactions on VLSI Systems best paper award, 2012 ACM/IEEE International Symposium on Low Power Electronics and Design best paper award, 2013 IEEE Transactions on VLSI Best paper award. Dr. Roy was a Purdue University Faculty Scholar (1998-2003). He was a Research Visionary Board Member of Motorola Labs (2002) and held the M. Gandhi Distinguished Visiting Faculty at the Indian Institute of Technology (Bombay) and Global Foundries visiting Chair at the National University of Singapore. He has been on the editorial board of IEEE Design and Test, IEEE Transactions on Circuits and Systems, IEEE Transactions on VLSI Systems, and IEEE Transactions on Electron Devices. He was Guest Editor for a Special Issue on Low-Power VLSI in the IEEE Design and Test (1994) and IEEE Transactions on VLSI Systems (June 2000), IEE Proceedings -- Computers and Digital Techniques (July 2002), and IEEE Journal on Emerging and Selected Topics in Circuits and Systems (2011). Dr. Roy is a fellow of IEEE.
\end{IEEEbiography}

\vfill
\end{document}